\documentclass[conference]{IEEEtran}
\IEEEoverridecommandlockouts
\usepackage{cite}
\usepackage{amsmath,amssymb,amsfonts}
\usepackage{algorithmic}
\usepackage{graphicx}
\usepackage{textcomp}
\usepackage{xcolor}
\usepackage{multirow}
\def\BibTeX{{\rm B\kern-.05em{\sc i\kern-.025em b}\kern-.08em
    T\kern-.1667em\lower.7ex\hbox{E}\kern-.125emX}}
\begin{document}

\title{PanelShield: Verifiable Closed-Loop Safe Planning for Robotic Industrial Panel Operation \\
\thanks{*Corresponding Author (Email: zhongxu\_hu@hust.edu.cn)}
}

\author{
\IEEEauthorblockN{Guipeng Xin, Jiahe Xu, Chenhui Wan, Jie Liu, Youmin Hu, and Zhongxu Hu*}
\IEEEauthorblockA{\textit{School of Mechanical Science and Engineering} \\
\textit{Huazhong University of Science and Technology} \\
Wuhan, China \\
\{xinguipeng, jiahe\_xu, wanchenhui, jie\_liu, youmhwh, zhongxu\_hu\}@hust.edu.cn}
}

\maketitle

\begin{abstract}
Industrial panel operation is knowledge-intensive and safety-critical. Beyond control recognition and action generation, execution must satisfy constraints in operation manuals and safety regulations. While foundation-model-based planners show strong semantic capability, they typically lack computable, localizable, and reproducible mechanisms for violation detection and repair. To address this, we propose \textit{PanelShield}, a verifiable closed-loop safety planning framework for manual-guided industrial panel operation. The framework generates parameterized action primitive sequences from \emph{task-relevant manual evidence} and applies dual formal verification with LTL and a Safety FSM to enforce cross-step temporal correctness and local transition legality. When violations occur, it outputs a structured counterexample with the earliest violating step and cause, enabling targeted repair and re-verification.

We build a multi-level long-horizon planning benchmark covering three representative industrial device panels, and evaluate the framework in simulation and real-world robotic experiments. Results show that PanelShield improves complex safety-constrained task performance over foundation-model-only planning baselines while reducing the violation rate to 2.7\%, with 4.1\,s total latency. Real-world experiments demonstrate end-to-end feasibility. Overall, PanelShield offers a verifiable approach to robotic panel operation that balances flexibility, safety, and auditability.
\end{abstract}


\section{Introduction}
Many critical devices in industrial production use operation panels as the primary human--machine interface (HMI). Operators monitor and control devices through knobs, buttons, and indicator signals, directly affecting process stability and safety \cite{mourtzis2023future}. In high-risk scenarios such as energized maintenance, fault recovery, and hazardous-area inspection, enabling robots to perform panel interaction has clear practical value. However, industrial panel operation is not merely a control-detection problem. Its core difficulty is the strong coupling between procedures and device states: the same control may require different action orders, allowable ranges, and interlock conditions under different modes or stages, while operations must satisfy strict preconditions, prohibitions, temporal dependencies, and threshold constraints specified in manuals and safety regulations \cite{ren2024embodied}. Because these constraints are scattered across long documents, raw-text planning context struggles to maintain long-horizon consistency and traceability, falling short of industrial requirements for \emph{verifiability}, \emph{auditability}, and \emph{correctability}.

\begin{figure}[t]
	\centering
	\includegraphics[clip, trim=0.0cm 0cm 0.0cm 0.0cm, width=\linewidth]{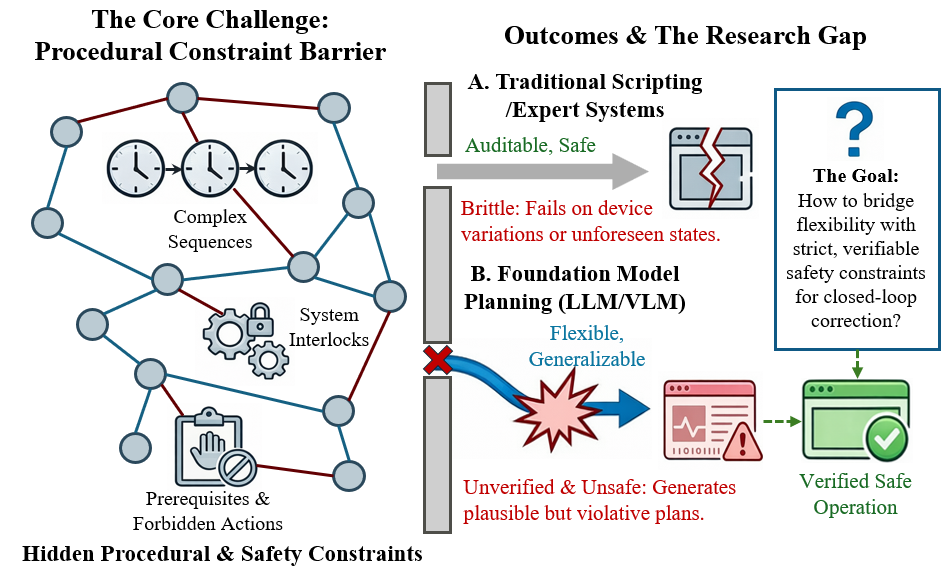}
        \vspace{-0.5cm}
        \setlength{\abovecaptionskip}{3pt}
	\caption{Core challenge of industrial panel operation. Dispersed procedures and safety constraints jointly form a procedural constraint barrier that is difficult to handle directly. Traditional script-based methods are safe but rigid, while foundation-model-based planning is flexible but prone to violations. Our goal is to preserve flexibility while enabling verifiable safety planning with closed-loop error correction.}
	\label{fig:teaser}
        \vspace{-0.7cm}
\end{figure}

Existing approaches face a trade-off. Expert workflows and rule scripts are auditable but tightly coupled to specific devices, leading to high migration and maintenance cost \cite{pan2012recent}. Foundation-model-based methods (e.g., VLM/LLM/VLA) offer semantic understanding and long-horizon reasoning \cite{liang2022code,huang2023voxposer,yang2025guiding,hu2023look,mon2025embodied}, but generated plans may violate interlocks, omit state checks, and fail to indicate \emph{which step} violates \emph{which rule}. A deployable system therefore needs computable procedural constraints and localizable feedback for closed-loop correction.

To address this, we propose \textit{PanelShield}, a verifiable closed-loop safety planning mechanism for manual-guided industrial panel operation. PanelShield generates a parameterized action-primitive sequence from \emph{task-relevant manual evidence}, constructs a computable execution trace, and performs dual formal verification: LTL checks global temporal correctness, while a Safety FSM checks local transition legality (e.g., modes and interlocks). When verification fails, the system outputs a localizable violation report with the rule identifier and earliest violating step, which is converted into a repair instruction for targeted revision and re-verification. This yields a counterexample-driven ``generate--verify--repair--re-verify'' loop, turning safety from post-hoc judgment into an executable pre-action gate.

The main contributions of this work are as follows:
\begin{enumerate}
    \item \textbf{Verifiable closed-loop safety planning paradigm:} We propose PanelShield, a plan-level safety assurance framework that grounds manual procedures into computable constraints.
    \item \textbf{Dual formal verification with localizable counterexamples:} We combine LTL and Safety FSM for complementary checking and produce rule- and step-level violation reports to enable targeted repair.
    \item \textbf{Validation on industrial long-horizon tasks:} We validate the method on a multi-level long-horizon benchmark and a real robotic arm platform, showing improved task success and reduced safety violations under acceptable latency.
\end{enumerate}

\begin{figure*}[t]
	\centering
	\includegraphics[width=0.95\textwidth]{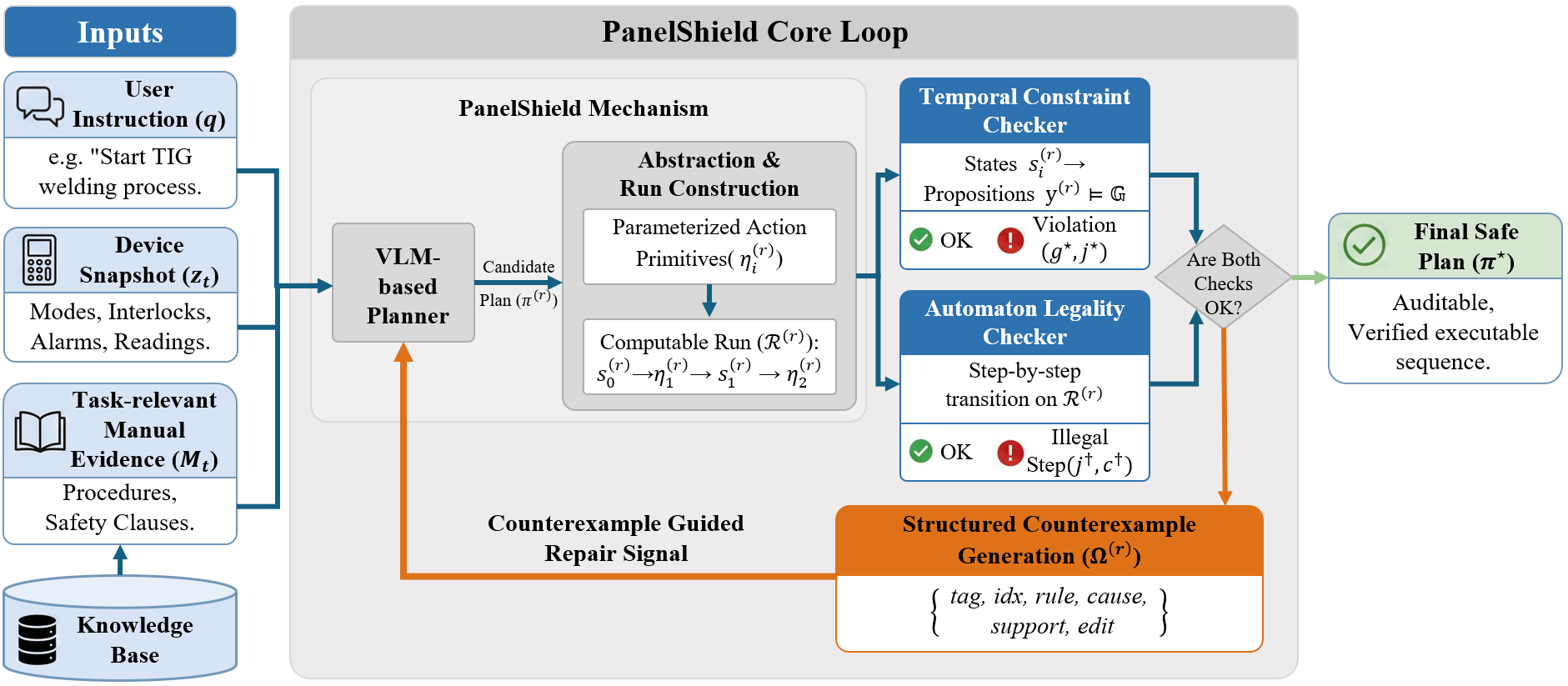}
    \vspace{-0.1cm}
    \setlength{\abovecaptionskip}{3pt}
	\caption{PanelShield core loop for verifiable closed-loop planning in industrial panel operation. A VLM-based planner generates a candidate plan from instruction, device state, and task-relevant manual evidence; the plan is then converted into a computable run and checked by complementary LTL-based temporal verification and Safety-FSM-based legality verification. Violations trigger structured counterexample generation and feedback-guided repair, and the loop repeats until a final safe executable plan is obtained.}
	\label{fig:method_full}
        \vspace{-0.5cm}
\end{figure*}

\section{Related Work}

\subsection{Foundation Models for Long-Horizon Planning}

VLM/LLM-driven high-level planning has lowered the barrier to task modeling by decomposing natural-language goals into multi-step action structures and showing semantic generalization in open environments \cite{fu2024can,shirai2024vision}. To improve long-horizon reliability, prior work has explored self-correction, structured plan representations, and task-and-motion planning (TAMP) integration \cite{wang2024llm,yang2025guiding,ao2025llm}.

However, industrial panel operation has \emph{strong procedural constraints} and \emph{strong state dependency}; the challenge is not only generating steps, but ensuring they are executable under rules. In generative frameworks, constraints are often prompt preferences or implicit priors, producing coherent but unverifiable plans. The central question is therefore: \emph{how to transform procedural knowledge from textual reminders into computable objects, so that plan compliance can be explicitly decided?}

\subsection{Task-Level Safety for LLM Agents}

Compared with motion-level collision risks, task-level safety issues more often stem from incorrect operation orders, unsatisfied preconditions, and violations of procedural prohibitions. Existing work on task-level safety for LLM agents can be roughly grouped into three directions: (i) formalizing safety requirements and checking them during planning/execution \cite{yang2024plug,wu2025selp,khan2025safety}; (ii) using safety critics or multi-model collaboration to evaluate and revise plans; and (iii) applying semantic filtering or refusal mechanisms before generation to block high-risk goals \cite{yin2024safeagentbench}. These approaches provide useful tools for injecting safety into LLM-based agents, but important gaps remain in industrial procedural settings.

First, many methods can identify that a plan is unsafe, but cannot reliably provide a \emph{localizable counterexample} (i.e., which step violates which rule), making targeted local repair difficult. Second, industrial panel operation requires both global cross-step temporal constraints and local legality checks for mode/interlock transitions, which are often not well covered by a single formalism.

\section{Methodology}

We propose \textit{PanelShield}, a formal-verification-based safety assurance mechanism that converts key procedural constraints in task-relevant manual evidence into computable LTL/FSM rules and produces a final executable safe plan through a closed-loop process. The rule base is compiled for each panel family from manuals and expert-audited templates; online planning only activates task-relevant clauses, reducing but not eliminating manual adaptation effort for new devices. Through dual checking and counterexample-guided repair, PanelShield captures procedural dependencies and transition boundaries, reducing risks from missing safety clauses or incorrect mode transitions while providing auditable and reproducible safety guarantees.

At each planning step, PanelShield takes a user instruction $q$, an observable device snapshot $z_t$ (e.g., mode bits, interlock flags, alarms, and key readings), and task-relevant manual evidence $M_t$ (procedure snippets and safety clauses relevant to the task), and outputs a verified safe plan $\pi^\star$. To avoid verification degenerating into re-interpreting natural-language plans, it represents each candidate plan as a sequence of parameterized action primitives and constructs a computable execution run for verification. It then performs two complementary checks: temporal constraint checking and automaton legality checking, and generates structured counterexamples when violations are detected to drive targeted repair.

\subsection{Evidence-Conditioned Planning}

A central difficulty in verifying industrial procedures is that natural-language plans lack stable decidable semantics, causing ``verification'' to rely on re-interpreting text with another model and making step-level localization unreliable. Formal safety verification, however, requires inputs with consistent parsable semantics. To directly bind violation localization and repair constraints to plan steps, PanelShield first generates a candidate plan conditioned on $(q, z_t, M_t)$:
\begin{equation}
\pi^{(0)} = \mathrm{Plan}(q, z_t, M_t).
\end{equation}

The candidate plan at the $r$-th iteration is uniformly represented as a sequence of parameterized action primitives:
\begin{equation}
\pi^{(r)} = \left[\eta_1^{(r)}, \eta_2^{(r)}, \dots, \eta_{T_r}^{(r)}\right],
\end{equation}
where
\begin{equation}
\eta_i^{(r)} = \left(u_i^{(r)}, \theta_i^{(r)}\right), \qquad u_i^{(r)} \in \mathcal{U}.
\end{equation}
Here, $\mathcal{U}$ denotes the primitive set (e.g., \texttt{mode-switch}, \texttt{safety-check}, \texttt{control-trigger}), and $\theta_i^{(r)}$ denotes primitive-specific parameters (e.g., target state/setpoint, direction, duration). This representation provides explicit step boundaries, which serve as anchors for stable back-references to the \emph{earliest violating step}.

To connect the action sequence and procedural constraints to a decidable carrier, we abstract the device snapshot $z_t$ into a discrete symbolic configuration $s \in \mathcal{S}$ and use a conservative or approximate transition operator to model state evolution after primitive execution:
\begin{align}
s_0^{(r)} &= \mathrm{Abs}(z_t), \\
s_{i+1}^{(r)} &= \mathrm{Trans}\!\left(s_i^{(r)}, \eta_{i+1}^{(r)}\right), \quad i=0,\dots,T_r-1.
\end{align}
We then construct the execution run
\begin{equation}
R^{(r)} = \left\langle s_0^{(r)}, \eta_1^{(r)}, s_1^{(r)}, \dots, \eta_{T_r}^{(r)}, s_{T_r}^{(r)} \right\rangle.
\end{equation}
Here, $\mathrm{Abs}(\cdot)$ maps observations to semantic quantities such as modes, interlocks, alarms, and key parameter intervals, while $\mathrm{Trans}(\cdot)$ provides consistent and decidable transition semantics. This run is not intended to certify continuous dynamics or perception correctness; ambiguous observations are conservatively treated as failed checks or trigger re-observation. It therefore supports procedural checking and violation localization while avoiding unsafe approval under uncertain symbolic states.

\subsection{Dual Formal Checking}

Safety requirements in industrial procedures can generally be divided into two types. The first type is \emph{global temporal properties} across steps, such as preconditions that must be satisfied before an operation, or operations that become forbidden after certain states occur. The second type is \emph{local transition legality} under modes/interlocks, such as actions disallowed under a given mode or transitions that must pass through intermediate states. To cover both, PanelShield executes a temporal checker and an automaton checker in parallel on $R^{(r)}$, respectively enforcing global temporal correctness and local transition legality, so that violations can be explicitly represented in the form of \emph{rule--location--cause}.

For global temporal properties, let $\mathcal{P}$ be the set of atomic propositions and define a labeling function
\begin{equation}
\Lambda : \mathcal{S} \rightarrow 2^{\mathcal{P}},
\end{equation}
which maps symbolic configurations in the run to proposition sets. This yields a proposition sequence
\begin{equation}
y^{(r)} = \left[\Lambda(s_0^{(r)}), \Lambda(s_1^{(r)}), \dots, \Lambda(s_{T_r}^{(r)})\right].
\end{equation}
Let $\mathcal{G}$ denote the set of global temporal constraints induced and compiled from task-relevant manual evidence. The temporal checker returns
\begin{equation}
\mathrm{TempCheck}\!\left(y^{(r)}, \mathcal{G}\right)=
\begin{cases}
\mathrm{OK}, & y^{(r)} \models \mathcal{G},\\
(g^\star, j^\star), & \text{otherwise},
\end{cases}
\end{equation}
where $g^\star$ identifies the violated constraint and $j^\star$ gives the earliest violation index, enabling direct formal localization of \emph{which rule} is violated and \emph{around which step} the violation occurs.

For local transition legality, we use a safety automaton $\mathcal{A}$ to represent allowable action triggers and transition structures under modes/interlocks, and check whether each step in $R^{(r)}$ is legal:
\begin{equation}
\mathrm{AutoCheck}\!\left(R^{(r)}, \mathcal{A}\right)=
\begin{cases}
\mathrm{OK}, & \text{if all steps are legal},\\
(j^\dagger, c^\dagger), & \text{otherwise},
\end{cases}
\end{equation}
where $j^\dagger$ returns the first illegal step and $c^\dagger$ provides a human-readable cause (e.g., \emph{missing intermediate mode} or \emph{action forbidden under interlock}), which directly supports subsequent repair. The dual-checking design gives the verification output a structured \emph{rule--location--cause} form and lays the foundation for closed-loop repair.

\subsection{Counterexample-Guided Repair Loop}

A one-shot verdict on a candidate plan is insufficient for knowledge-intensive long-horizon tasks. When a violation occurs, the system must convert verification results into actionable repair signals so that replanning becomes a constrained local modification process rather than unconstrained rewriting. To this end, PanelShield uniformly packages failures from the temporal checker and automaton checker into a structured counterexample and uses it to constrain the next planning iteration, improving convergence and plan stability through minimal local edits.

When either $\mathrm{TempCheck}$ or $\mathrm{AutoCheck}$ fails, PanelShield merges the failure outputs into a structured counterexample
\begin{equation}
\Omega^{(r)} =
\{\texttt{tag},\texttt{idx},\texttt{rule},\texttt{cause},\texttt{support},\texttt{edit}\},
\end{equation}
where \texttt{idx} points to the earliest violating step, \texttt{rule} corresponds to $g^\star$ or an automaton constraint, \texttt{cause} is the explanation, \texttt{support} links back to the relevant clause fragments in $M_t$, and \texttt{edit} specifies the direction of minimal local modification (e.g., inserting a required check/transition, reordering steps, or replacing an illegal primitive).

Conditioned on the counterexample, the planner generates the next candidate plan:
\begin{equation}
\pi^{(r+1)} = \mathrm{Plan}(q, z_t, M_t, \Omega^{(r)}).
\end{equation}
PanelShield then repeats the loop of \emph{run construction $\rightarrow$ dual checking $\rightarrow$ counterexample generation $\rightarrow$ targeted repair} until both checks return $\mathrm{OK}$ or the iteration budget is reached, and finally outputs $\pi^\star$.

Through this counterexample-guided closed loop, PanelShield upgrades task-level safety from one-shot prompting/filtering to a localizable, reproducible, and convergent error-correction mechanism, thereby reducing high-risk execution errors caused by incorrect task ordering, missing preconditions, or illegal transitions.

\section{Experiments}

\subsection{Benchmark Setup}

For evaluation, we build an industrial panel operation planning benchmark covering three representative device panels: a Variable Frequency Drive (VFD) controller, a Power Management System, and a Hydraulic Control Unit. For each device category, we provide a complete operation manual, safety regulations, and technical specification documents. To evaluate retrieval and planning under realistic document conditions, we additionally define \emph{task-relevant manual evidence}: for each task instruction, we annotate the document content directly relevant to planning, and use it to assess the consistency between retrieved evidence and generated plans.

To reflect task complexity, we design a three-level natural-language instruction set: \textbf{Level-1} (single-step adjustment), \textbf{Level-2} (multi-step sequential operation), and \textbf{Level-3} (composite tasks with explicit safety constraints). The final benchmark contains 300 rigorously validated user instructions, each paired with an expert-defined reference task plan sequence as ground truth.

For real-world experiments, we use a motor drive controller simulator as the physical target device, which provides observable responses and state feedback for typical control commands. The robotic system is implemented with a ROS~2 architecture and includes three modules: visual perception, task planning, and motion control. The perception module uses a RealSense D435i depth camera for control detection and pose estimation. The planning module receives user instructions and generates task plans using online-retrieved safety constraints and operational evidence. The motion control module uses the official UR3 controller for trajectory planning and precise interaction with target controls.

In simulation, \textbf{Success Rate} measures task completion within a maximum reasoning budget, while \textbf{Violation Rate} / \textbf{Compliance Rate} assess execution safety. We also report average repair iterations (\textbf{Repair}) and runtime overhead (\textbf{Latency}) to evaluate convergence and online usability. Real-world experiments validate end-to-end deployability, focusing on reasoning and execution time in chained tasks.

\begin{figure}[t]
	\centering
	\includegraphics[clip, trim=0.0cm 0cm 0.0cm 0.0cm, width=\linewidth]{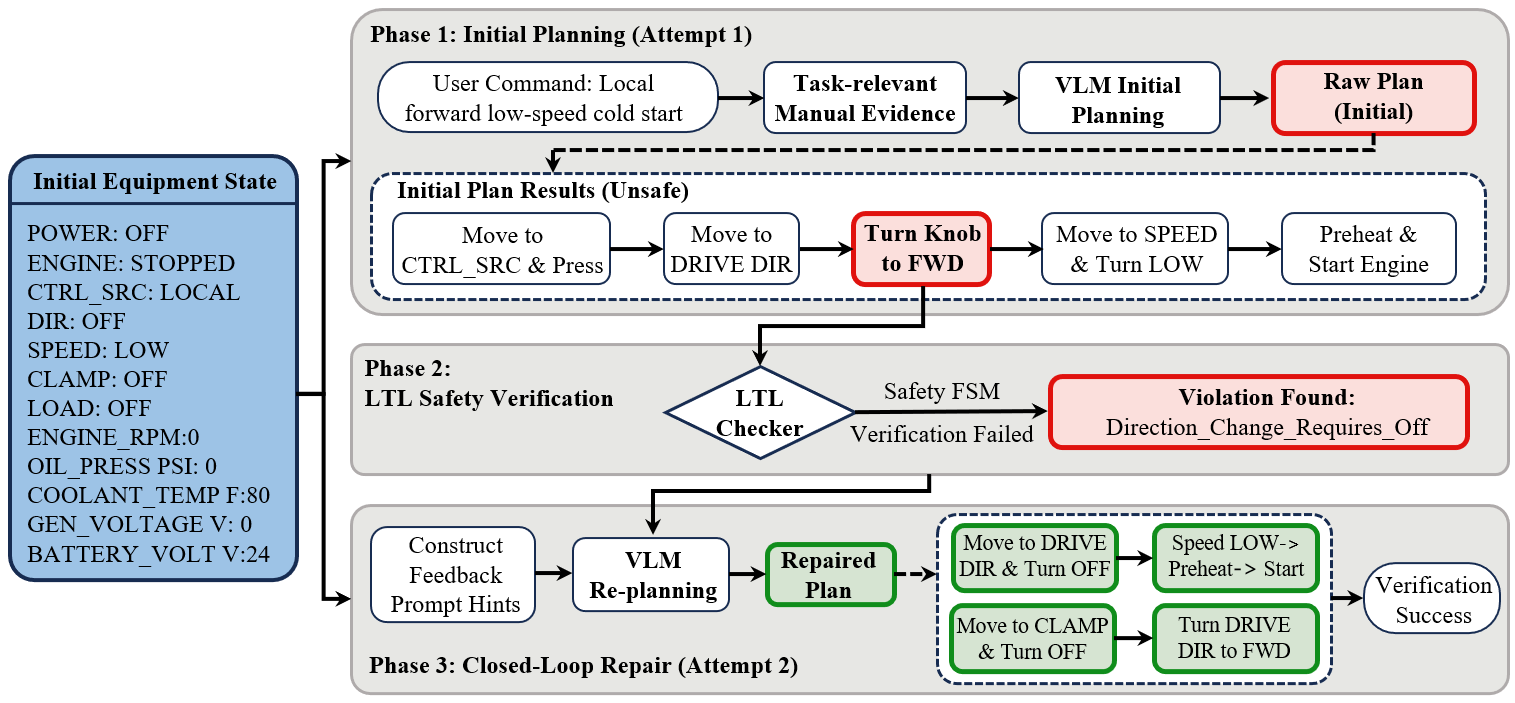}
        \vspace{-0.5cm}
        \setlength{\abovecaptionskip}{3pt}
	\caption{Example of counterexample-guided closed-loop repair in PanelShield.}
	\label{fig:Example}
        \vspace{-0.7cm}
\end{figure}

\subsection{Experimental Results Analysis}

\subsubsection{Simulation Planning Baselines}

As illustrated in Fig.~\ref{fig:Example}, the case study uses the instruction \emph{``low-speed forward cold start in local mode''}. The system first retrieves task-relevant \emph{manual evidence}, reads the current device state, and generates an initial task sequence with a VLM. Before execution, LTL checking and Safety FSM verification detect a violation of the rule \emph{``direction switching must return to OFF first''}, so the plan is rejected.
The system then enters a closed-loop repair stage: it feeds back the violation cause and relevant rules to the VLM, re-generates the plan, and re-verifies it until a safe plan is obtained. 

\begin{figure}[t]
	\centering
	\includegraphics[clip, trim=0.0cm 0cm 0.0cm 0.0cm, width=\linewidth]{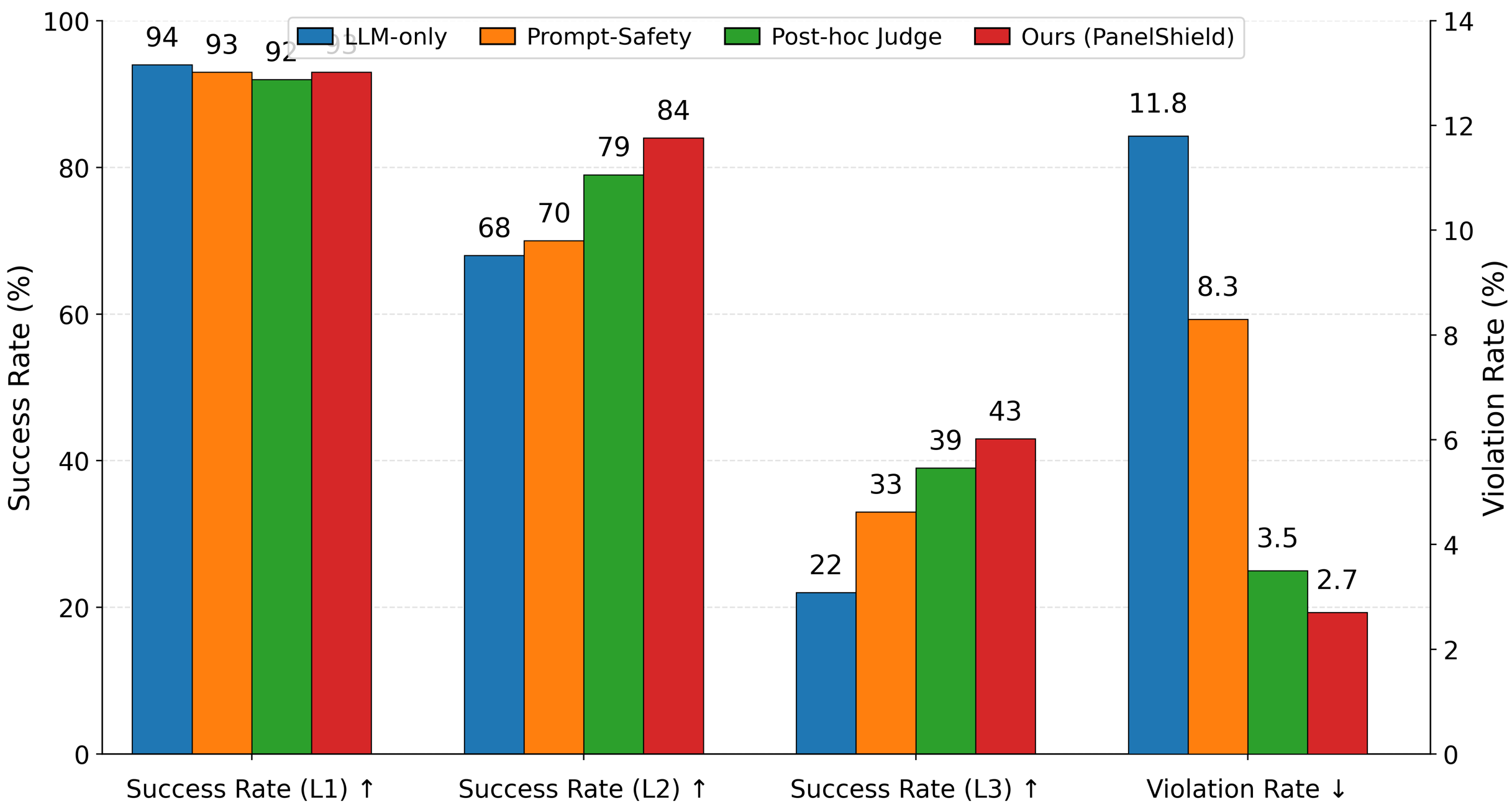}
        \vspace{-0.5cm}
        \setlength{\abovecaptionskip}{3pt}
	\caption{Simulation planning baseline results on industrial panel operation tasks.}
	\label{fig:plan}
        \vspace{-0.7cm}
\end{figure}

Fig.~\ref{fig:plan} reports the simulation planning baseline results. We compare four representative methods: \textbf{LLM-only}, which directly generates and executes an action sequence from the instruction and manual context; \textbf{Prompt-Safety}, which encodes key procedural rules as soft prompt constraints; \textbf{Post-hoc Judge}, which checks compliance after planning using an LLM/rule-based reviewer; and \textbf{Ours (PanelShield)}, which uses dual checking with LTL and a Safety FSM, and outputs the earliest violating step and cause to drive targeted repair.

PanelShield achieves the best overall performance on complex-task success and safety. Its success rates on L1/L2/L3 are \textbf{93.0\% / 84.0\% / 43.0\%}. In particular, on \textbf{L3} tasks, it reaches the highest success rate (\textbf{43.0\%}), improving over LLM-only (22.0\%), Prompt-Safety (33.0\%), and Post-hoc Judge (39.0\%) by \textbf{21.0}, \textbf{10.0}, and \textbf{4.0} percentage points, respectively. It also achieves the lowest violation rate (\textbf{2.7\%}), outperforming Prompt-Safety (8.3\%) and Post-hoc Judge (3.5\%). This residual rate is not acceptable for unattended safety-critical deployment, but it highlights the need to further strengthen rule coverage, perception grounding, and repair robustness. Overall, computable constraints with counterexample-guided repair improve success while reducing violations in complex panel planning.

\begin{table}[h]
    \centering
    \vspace{-0.3cm}
    \caption{Repair iterations and planning time overhead results.}
    \renewcommand{\arraystretch}{1.3}
    \resizebox{0.49\textwidth}{!}{%
    \begin{tabular}{c|cccc}
    \hline
         \multirow{2}{*}{Method} & Repair$\downarrow$ & Total Latency$\downarrow$ & Planning Time$\downarrow$ & Verification Time$\downarrow$ \\
         ~ & (avg. iters) & (s) & (s) & (s) \\
        \hline
        LLM-only & - & 1.7 & 1.7 & - \\
        Prompt-Safety & - & 2.2 & 2.2 & - \\
        Post-hoc Judge & 3.9 & 4.5 & 3.4 & 1.1 \\
        \textbf{Ours (PanelShield)} & \textbf{3.6} & \textbf{4.1} & 3.5 & \textbf{0.6} \\
        \hline
    \end{tabular}}
    \label{tab:repair_latency_results}
    \vspace{-0.3cm}
\end{table}

\subsubsection{Repair Iterations and Planning Time Overhead}

As shown in Table~\ref{tab:repair_latency_results}, only \textbf{Post-hoc Judge} and \textbf{our method} provide explicit repair. Post-hoc Judge requires more iterations because, without the \emph{earliest violating step} and \emph{violation cause}, it often rewrites the whole plan. In contrast, our LTL and Safety FSM checks produce structured counterexamples and restrict repair to local revisions near the violation, yielding fewer iterations and more stable convergence.

For time overhead, \textbf{LLM-only} and \textbf{Prompt-Safety} are fastest because they omit formal verification. \textbf{Post-hoc Judge} is slowest due to extra review and repeated rewrites. Although our method introduces dual verification (LTL + Safety FSM), the checkers are lightweight and faster than LLM-based judging, keeping latency acceptable while improving safety.

Overall, the computation cost of formal verification is controllable, and structured counterexample-guided repair reduces unnecessary regeneration, yielding a better safety trade-off.

\begin{table}[h]
    \centering
    \vspace{-0.3cm}
    \caption{Ablation results on violation localization, violation decomposition, and repair locality.}
    \renewcommand{\arraystretch}{1.3}
    \resizebox{0.49\textwidth}{!}{%
    \begin{tabular}{c|cccc}
    \hline
         \multirow{2}{*}{Method} & Earliest-violation Step$\downarrow$ & LTL Violations$\downarrow$ & FSM Violations$\downarrow$ & Avg Edit Size / Repair$\downarrow$ \\
         ~ & (steps) & (\%) & (\%) & (steps) \\
        \hline
        Full (LTL + FSM + Repair) & \textbf{2.1} & \textbf{2.0} & \textbf{0.7} & \textbf{1.3} \\
        w/o LTL (FSM + Repair) & 2.4 & - & 1.6 & 1.8 \\
        w/o FSM (LTL + Repair) & 2.5 & 2.8 & - & 1.9 \\
        w/o Repair (LTL + FSM, no repair) & 2.0 & 2.1 & 0.8 & - \\
        Post-hoc Judge & 3.6 & 2.4 & 1.1 & 3.0 \\
        \hline
    \end{tabular}}
    \label{tab:ablation_verification_repair}
    \vspace{-0.3cm}
\end{table}

\subsubsection{Ablation on Verification and Repair Modules}

Table~\ref{tab:ablation_verification_repair} presents an ablation study of the verification and repair modules from three aspects: violation localization, violation-source decomposition, and repair locality. \emph{Earliest-violation Step} denotes the first action index where a formal constraint violation is triggered, reflecting how early the system can expose errors in the execution chain. \emph{LTL/FSM Violations (\%)} report the proportions of violations detected by LTL and Safety FSM, respectively. \emph{Avg Edit Size / Repair} measures the average number of action-level edits per repair iteration.

The full method (\textbf{Full}) achieves the lowest violation rates under both checkers, showing that joint LTL + Safety FSM verification can constrain complementary sources of unsafe plans. Removing either verifier increases the burden on the other and leads to higher violation rates. For example, the \emph{Earliest-violation Step} rises to 2.4 and 2.5 for \emph{w/o LTL} and \emph{w/o FSM}, indicating weaker ability to consistently truncate unsafe plans at early stages.

Meanwhile, \emph{w/o Repair} can still detect violations relatively early through dual verification, but cannot form a \emph{localization--repair--verification} closed loop because counterexample-guided repair is disabled. \emph{Post-hoc Judge} shows a much later \emph{Earliest-violation Step} and the largest average repair edit size, indicating a tendency toward late holistic review and plan rewriting rather than targeted local repair.

In contrast, the full method yields the smallest average edit size, consistent with the repair-iteration and latency results, confirming that structured localization improves convergence and repair efficiency.

\begin{figure}[t]
	\centering
	\includegraphics[clip, trim=0.0cm 0cm 0.0cm 0.0cm, width=\linewidth]{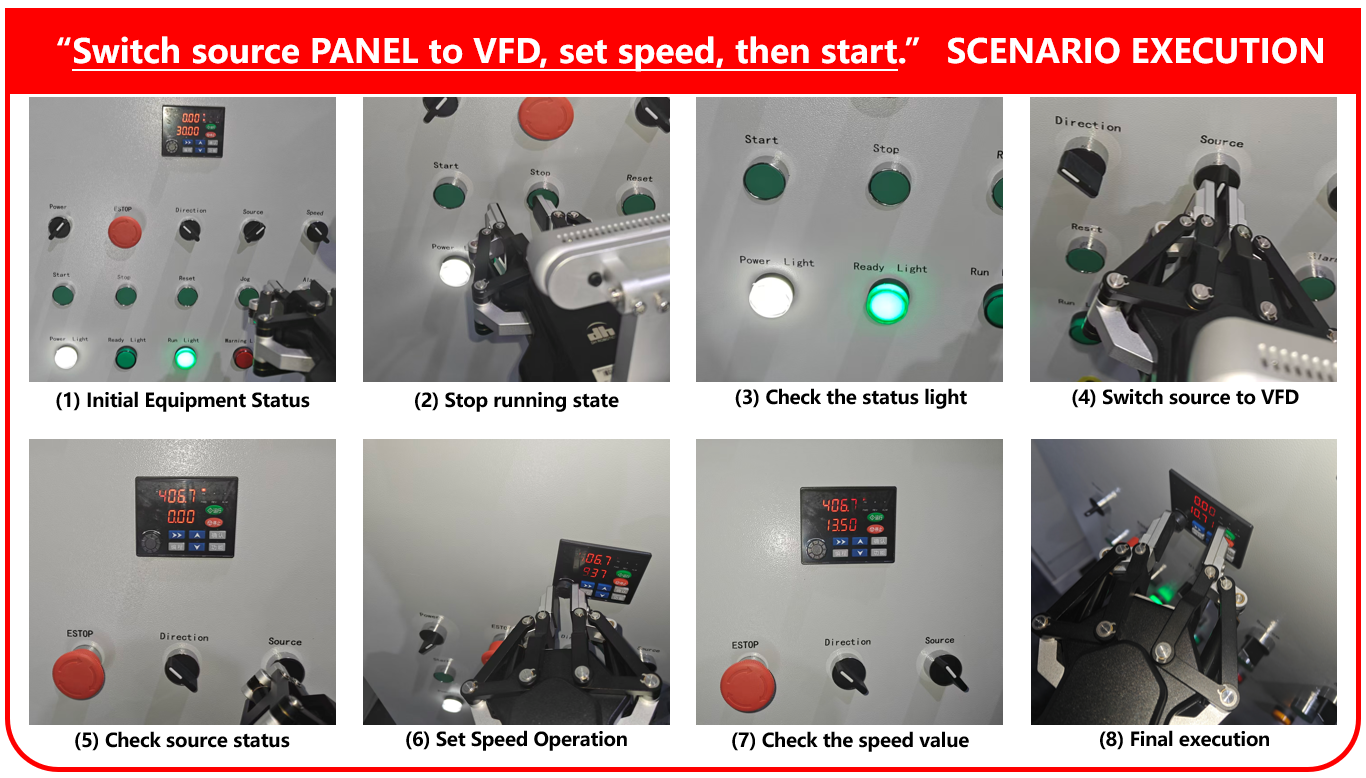}
        \vspace{-0.5cm}
        \setlength{\abovecaptionskip}{3pt}
	\caption{Real-world chained-task execution example.}
	\label{fig:real}
        \vspace{-0.7cm}
\end{figure}

\subsubsection{Real-World End-to-End Experiments}

In real-world experiments, we evaluate the end-to-end deployability of the proposed method on an actual device, focusing on execution efficiency and long-horizon stability. Before critical transitions, state feedback is re-read; ambiguous readings are treated as failed checks rather than trusted inputs. As reported in Table~\ref{tab:real_chain_results}, inference time is 3.5--5.6\,s and execution time is 4.2--10.7\,s using GPT-4o as the planner, indicating practical responsiveness for chained execution.

As shown in Fig.~\ref{fig:real}, in the task \emph{``Switch source from PANEL to VFD, set speed, then start''}, the system demonstrates effective task understanding and decomposition. It identifies the required operation steps, determines execution conditions from the current device state, reuses operation parameters from the previous task, and performs subtask-level checks on critical states. This verifies key states at each stage and helps avoid operation errors.

\begin{table}[h]
    \centering
    \vspace{-0.3cm}
    \caption{Real-world results on long-horizon chained tasks.}
    \renewcommand{\arraystretch}{1.3}
    \resizebox{0.48\textwidth}{!}{%
    \begin{tabular}{c|cc}
    \hline
         \multirow{2}{*}{Task} & \multirow{2}{*}{Inference Time (s)$\downarrow$} & \multirow{2}{*}{Execution Time (s)$\downarrow$} \\
         ~ & ~ & ~ \\
        \hline
        1. Turn on main power & 3.5 & 5.8 \\
        2. Switch low speed under PANEL & 4.2 & 5.8 \\
        3. Change the direction to REV & 3.8 & 6.1 \\
        4. Perform a jog operation & 3.7 & 4.2 \\
        5. Verify READY, then START & 5.6 & 10.7 \\
        \hline
    \end{tabular}}
    \label{tab:real_chain_results}
    \vspace{-0.3cm}
\end{table}

\section{Conclusion}

To address key challenges in industrial panel operation, including dispersed procedural knowledge, strong state coupling, and high error cost, we propose \textit{PanelShield}, a verifiable closed-loop safety planning framework, and validate it in both simulation and real-world robotic execution. Results show that the framework largely preserves performance on low-complexity tasks while significantly improving success rates on high-complexity tasks and reducing violation rates. Real-world case studies further demonstrate that the method can generate procedure-compliant operation sequences and verify critical steps through feedback such as indicator lights and display values, enabling stable chained-task execution on real devices.

For deployment in more complex industrial scenarios, three limitations remain. First, new panel families still require expert-audited compilation of manual clauses into LTL/FSM templates; future work should automate evidence extraction while retaining human approval for safety rules. Second, verification is only as reliable as the symbolic state derived from perception, so confidence thresholds, repeated observation, and fail-safe stops are needed under noise or localization errors. Third, residual violations may arise from incomplete rule coverage, planner hallucination, or physical/perception mismatch; in safety-critical operation these cases should block execution rather than be treated as acceptable errors.

\section*{Acknowledgment}

This work was supported in part by the National Major Science and Technology Projects of China (No.2025ZD1606204), National Natural Science Foundation of China (No.52575572, No.52575115), National Key R\&D Program of China (2024YFE03140003),  Science and Technology Program of Hubei Province (2024BEB025), Joint Research Project of the Yangtze River Delta Science and Technology Innovation Community (2024CSJGG1400), UNSW-HUST Global Research \& Impact Program (5003100202).

\bibliography{references}

\end{document}